\documentclass[lettersize,journal]{IEEEtran}
\usepackage{amsmath,amsfonts}
\usepackage{algorithmic}
\usepackage{algorithm}
\usepackage{array}
\usepackage[caption=false,font=normalsize,labelfont=sf,textfont=sf]{subfig}
\usepackage{textcomp}
\usepackage{stfloats}
\usepackage{xcolor} 
\usepackage{url}
\usepackage{verbatim}
\usepackage{graphicx}
\usepackage{enumitem}
\usepackage{multirow}
\usepackage{cite}
\usepackage{balance}
\begin{document}

\title{Federated Cross-Training Learners for Robust Generalization under Data Heterogeneity}

\author{Zhuang Qi,
        Lei Meng\IEEEauthorrefmark{1},~\IEEEmembership{Member,~IEEE}\thanks{Lei Meng is the corresponding author.},
        Ruohan Zhang,
        Yu Wang,
        Xin Qi,
        Xiangxu Meng, \\
        Han Yu, \IEEEmembership{Senior~Member,~IEEE},
        Qiang Yang, \IEEEmembership{Fellow,~IEEE}

\thanks{Zhuang Qi is with School of Software, Shandong University, China. Email: z\_qi@mail.sdu.edu.cn.}

\thanks{Lei Meng is with Shandong University and Shandong Research Institute of Industrial Technology, China. Email: lmeng@sdu.edu.cn.}

\thanks{Ruohan Zhang is with Inspur, China. Email: zhangrh01@inspur.com.}

\thanks{Yu Wang is with Shandong Research Institute of Industrial Technology. Email: wangyu@sriit.cn.}

\thanks{Xin Qi is with School of Chemistry and Life Sciences, Suzhou University of Science and Technology, China. Email: qixin@usts.edu.cn.}

\thanks{Xiangxu Meng is with School of Software, Shandong University, China. Email: mxx@sdu.edu.cn.}

\thanks{Han Yu is with College of Computing and Data Science, Nanyang Technological University. Email: han.yu@ntu.edu.sg.}

\thanks{Qiang Yang is with the PolyU Academy for Artificial Intelligence, Hong Kong Polytechnic University, Hong Kong. Email: profqiang.yang@polyu.edu.hk.}
}

\markboth{Journal of \LaTeX\ Class Files,~Vol.~14, No.~8, August~2021}%
{Shell \MakeLowercase{\textit{et al.}}: A Sample Article Using IEEEtran.cls for IEEE Journals}


\maketitle

\begin{abstract}



Federated learning benefits from cross-training strategies, which enables models to train on data from distinct sources to improve generalization capability. However, due to inherent differences in data distributions, the optimization goals of local models remain misaligned, and this mismatch continues to manifest as feature space heterogeneity even after cross-training. We argue that knowledge distillation from the personalized view preserves client-specific characteristics and expands the local knowledge base, while distillation from the global view provides consistent semantic anchors that facilitate feature alignment across clients. To achieve this goal, this paper presents a cross-training scheme, termed FedCT, includes three main modules, where the consistency-aware knowledge broadcasting module aims to optimize model assignment strategies, which enhances collaborative advantages between clients and achieves an efficient federated learning process. The multi-view knowledge-guided representation learning module leverages fused prototypical knowledge from both global and local views to enhance the preservation of local knowledge before and after model exchange, as well as to ensure consistency between local and global knowledge. The mixup-based feature augmentation module aggregates rich information to further increase the diversity of feature spaces, which enables the model to better discriminate complex samples. Extensive experiments were conducted on four datasets in terms of performance comparison, ablation study, in-depth analysis and case study. The results demonstrated that FedCT alleviates knowledge forgetting from both local and global views, which enables it outperform state-of-the-art methods.




\end{abstract}

\begin{IEEEkeywords}
Federated learning, Cross-training, Knowledge forgetting, Prototypical distillation, Non-IID data
\end{IEEEkeywords}

\vspace{-0.2cm}
\section{Introduction}\label{}
\IEEEPARstart{F}ederated learning (FL) aims to address the challenge of data privacy leakage posed by traditional centralized machine learning methods in scenarios where data is distributed across multiple devices or organizations \cite{yuan2024bm,mcmahan2017communication,miao2023fedseg,qi2024attentive}. It enables collaborative model training across distributed devices or data sources while respecting data privacy and security \cite{kairouz2021advances,fu2025Alignments,liao2025Vertical,jiang2024data}. Existing studies leverage model-level interactions and aggregation between participants and the server to achieve updates of the global model, which facilitates each participant to benefit from the training knowledge of others and collectively enhance the generalization of the global model \cite{huang2025Coordinator,fu2023Unified,hu2024kdd,meng2024improving}. However, recent studies have revealed the fact that federated learning encounters a decline in the performance of the global model when confronted with variations in data distribution among different data sources \cite{hu2024fedmut,wang2024aggregation,lu2024federated,qi2023cross,cai2024lgfgad}. This can be largely attributed to label distribution skew, as it causes clients to optimize for different local objectives. As a result, this misalignment in optimization leads to inconsistencies in the learned feature representations, which is commonly known as feature space heterogeneity \cite{qi2023cross,mu2023fedproc}.


\begin{figure}[t]
\centering
\includegraphics[width=0.45\textwidth]{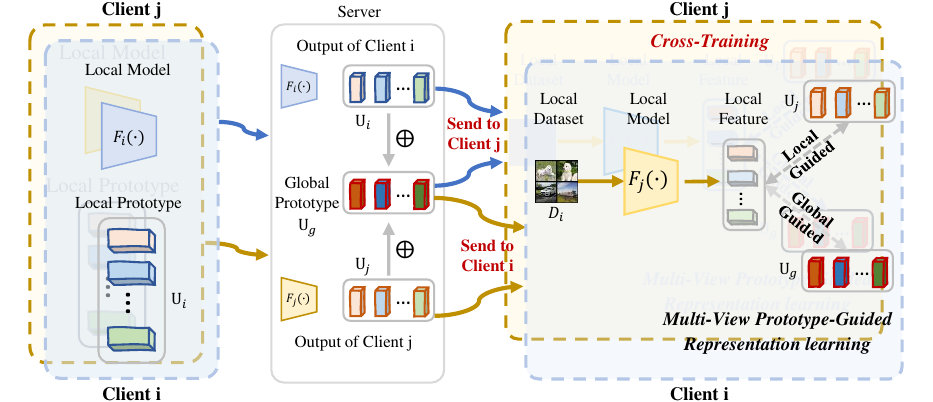}
\caption{Motivation of FedCT. It integrates personalized and global knowledge to jointly harness their strengths, which simultaneously expanding the trainable sample set of local models while mitigating the inconsistency issue between local and global optimization objectives. FedCT enhances the preservation of local knowledge before and after cross training, as well as the consistency between local and global knowledge.} 
\label{fig:mg}
\end{figure}

To mitigate the issue of feature heterogeneity, existing methods typically focus on improving the performance of local models on multi-source datasets, rather than on a single data source. They either integrate cross-training strategies to assist local models in training on multi-source datasets \cite{mao2020fedexg,matsuda2022fedme,liucross}, which can expand the knowledge base that local models can learn from, or use external knowledge to guide the training of local models. This adjustment in their optimization objectives enhances the cooperative modeling capabilities among local models \cite{li2021model,mu2023fedproc,wei2022knowledge}. For example, FedExg is recognized as the first cross-training method, which achieves an impressive performance \cite{mao2020fedexg}. Moreover, FedMe employs mutual learning techniques to enable client models to acquire knowledge from other data sources \cite{matsuda2022fedme}. And PGCT utilizes prototypes to guide local models in learning a consistent feature space across different clients, which alleviates the issue of knowledge forgetting in cross-training processes \cite{liucross}. 

\begin{figure}
\centering
\includegraphics[width=0.5\textwidth]{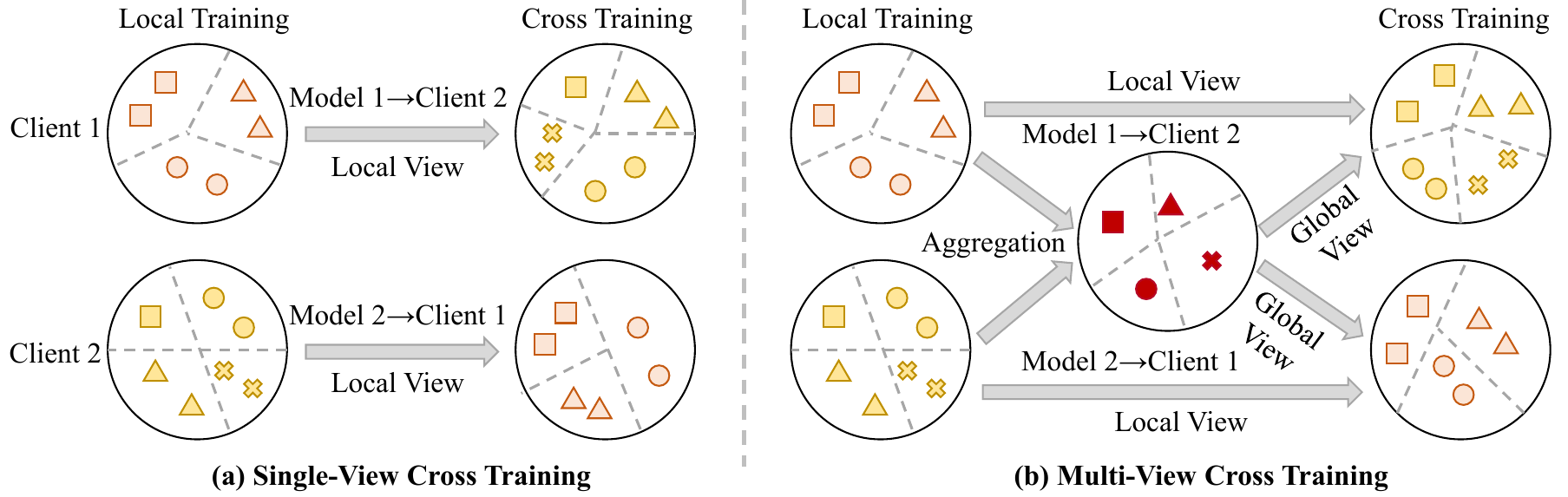}
\caption{(a) Feature alignment confined to exchange groups under local-view cross-training leads to persistent inter-client representation heterogeneity. (b) Incorporating global-view knowledge distillation helps mitigate feature space heterogeneity by simultaneously promoting local knowledge retention and global representation alignment.}

\label{fig11}
\end{figure}

\begin{figure}
\centering
\includegraphics[width=0.5\textwidth]{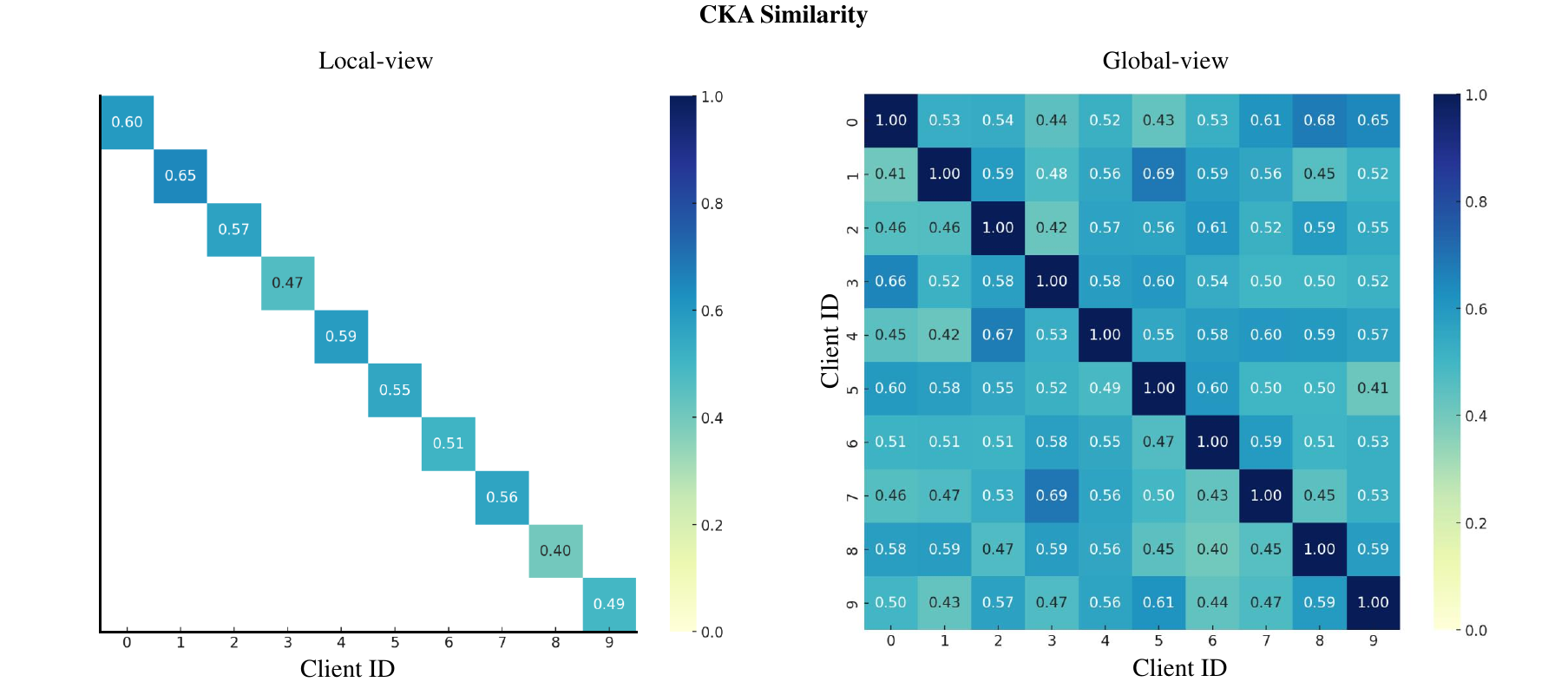}
\caption{
Multi-view knowledge forgetting in traditional cross training. The local view shows the centered kernel alignment (CKA) similarity of features from a local model before and after cross training, while the global view reflects the CKA similarity among features from all local models.
} 
\label{fig12}
\end{figure}

Despite yielding performance gains, they primarily focus on alleviating knowledge forgetting by ensuring consistency within local exchange groups. Specifically, they emphasize whether the knowledge retained by a client remains stable after interacting with other clients in a limited subset. However, such groups typically comprise only a small fraction of the entire federation, which is insufficient to expose clients to the full diversity of global feature representations. As depicted in Figure \ref{fig11}(a), the model 1(or 2) primarily focuses on replaying its original local knowledge when trained on client 2(or 1), but fails to account for global feature consistency. As a result, feature space heterogeneity across clients persists after cross training. To bridge this gap, we propose to move beyond purely local consistency and instead facilitate feature alignment from both local and global views. In particular, the local view expands each model’s knowledge base with semantically relevant information and retains prior knowledge to mitigate forgetting during cross training, while the global view offers consistent semantic anchors that align representations across clients. By integrating both, clients are exposed to complementary information that helps reduce representational bias and promotes latent space consistency across heterogeneous data sources, as shown in Figure~\ref{fig11}(b) and Figure \ref{fig12}.

Based on the above analysis, we propose a multi-view knowledge-guided cross-training framework, termed FedCT, to tackle the challenge of feature space heterogeneity in federated learning. As illustrated in Figure~\ref{fig:mg}, FedCT simultaneously expands each client’s local knowledge base and aligns their representations with global semantics by integrating information from both the local view and the global view. 
To this end, FedCT consists of three key modules: 1) the Consistency-Aware Knowledge Broadcasting module assigns each local model to a suitable exchange group based on its compatibility with shared class-wise prototypes. These prototypes serve as abstract semantic anchors that summarize intra-class knowledge. This encourages knowledge exchange among semantically aligned clients, promoting more effective collaboration; 2) the Multi-View Knowledge-Guided Representation Learning module enriches the local knowledge base and guides clients in constructing as homogeneous as possible class-level representations in the latent space to alleviate the challenge of knowledge forgetting from multiple views, which facilitates a more comprehensive understanding of global distributions and inter-class relationships; 3) the Mixup-based Feature Augmentation module focuses on enhancing the decision phase, and it aggregates rich information from both prototypes and features, which increases the diversity of the feature space. This reduces the risk of overfitting and contributes to improving the generalization performance and robustness of local models.

Extensive experiments are conducted on four commonly used datasets in terms of performance comparison, ablation study of the key components of FedCT, and case study for the effectiveness of knowledge broadcasting and representation learning. The results demonstrate that multi-view knowledge-guided cross-training can expand the learnable knowledge of local models and alleviate dataset bias from the perspective of representation consistency. To summarize, this paper makes three main contributions:
\begin{itemize}[leftmargin=10pt]
    \item This paper presents a model-agnostic cross-training strategy, named FedCT, to mitigate the problem of knowledge forgetting caused by dataset bias. It integrates multi-view knowledge to preserve client-specific semantics while promoting global consistency, enhancing representation robustness under heterogeneous data distributions. 
    
    
    \item This paper presents a plug-and-play technique, modeling the cooperative relationship between clients based on the prototypical prediction loss, that can enhance collaboration benefits between clients in scenarios with heterogeneous data distributions and can be easily incorporated into existing methods to improve communication efficiency. 
    
    \item Experimental findings have revealed that local and global knowledge forgetting diminishes the benefits that cross-training strategies bring to federated learning. These findings ensure the effectiveness of the proposed multi-view knowledge-guided representation learning module, which mitigates the issue of multi-view knowledge forgetting.
  

\end{itemize}

\section{Related Work}\label{}



\subsection{Methods for Local Training Optimization.}
To mitigate dataset biases, extensive research has focused on client-side solutions, particularly regularization-based methods and cross-training strategies. Both aim to promote consistent knowledge learning across clients. Regularization-based methods introduce carefully designed constraints from three main perspectives: model parameters or gradients \cite{gao2022feddc,li2020federated,luo2023gradma,ruan2022fedsoft,shoham2019overcoming}, feature space \cite{gong2021ensemble,li2021model,michieli2021prototype,mu2023fedproc,shen2020federated,shi2022towards,tan2022fedproto,ye2022fedfm,yu2021fed2,zhang2021federated,seo2024relaxed}, and model predictions \cite{lee2022preservation,zhang2022federated,zhu2021data}. Weight-based regularization is commonly used to prevent divergence by keeping local model parameters close to the global model. For instance, FedProx employs a Euclidean distance penalty \cite{li2020federated}, while FedDC adds a correction term to address parameter gaps \cite{gao2022feddc}.
Feature-based regularization focuses on aligning client representations. MOON introduces a model-level contrastive loss \cite{li2021model}, and methods like FedProc \cite{mu2023fedproc} and FPL \cite{huang2023rethinking} use class-aware prototypical contrast to guide local learning.
Prediction-based regularization uses the global model’s output probabilities as soft targets. FedNTD \cite{lee2022preservation} and FCCL++ \cite{huang2023generalizable} apply distillation on non-true classes to reduce forgetting.

Moreover, cross-training has emerged as a promising strategy to mitigate dataset bias. It enables model exchange in a privacy-preserving manner, allowing local models to be re-trained across clients and thus exposed to more diverse data \cite{liucross,mao2020fedexg,matsuda2022fedme}. This expands the local training set and facilitates more comprehensive knowledge acquisition. 
Importantly, cross-training is orthogonal to regularization methods and can be combined to achieve further performance improvements. For example, PGCT \cite{liucross} addresses local knowledge forgetting by leveraging prototypes to guide clients toward consistent decision boundaries. While these methods expand local knowledge, ensuring cross-client knowledge consistency remains a key challenge, as illustrated in Figure~\ref{fig11}.


\subsection{Methods for Improving Aggregation on the Server}


The aggregation phase plays a critical role in federated learning, as it determines how local updates are integrated into the global model. Existing approaches to improve aggregation efficacy can be grouped into three categories: (1) designing alternative rules to replace standard weighted averaging \cite{chen2023elastic,chen2020fedbe,pillutla2022robust,wang2020federated,wang2020tackling,yuan2023collaborative}, (2) fine-tuning or retraining after aggregation \cite{luo2021no,pi2023dynafed,shang2022federated,wang2023fedftha,zhang2022fine}, and (3) clustering-based client grouping \cite{duan2021flexible,li2022data,long2023multi,wei2023edge}. For instance, FedNova applies normalization before aggregation to reduce target inconsistency and accelerate convergence \cite{wang2020tackling}, while FedBE samples high-quality models and combines them using Bayesian model averaging \cite{chen2020fedbe}. Hu. et al \cite{hu2024aggregation} present an interesting model recombination paradigm to effectively guide the FL training towards a flatter area, then improve the generalization of the model.

In addition, fine-tuning or retraining the model after aggregation has also demonstrated improved performance. It aims to integrate knowledge from multiple data sources to construct a stable global model. For example, FedFTG first models the input space of multiple parties to generate pseudo-data, and then utilizes knowledge distillation to correct the biases introduced after model aggregation \cite{zhang2022fine}. CCVR \cite{luo2021no} and CReFF \cite{shang2022federated} believe that the biases in the model primarily stem from the classifier, and therefore, it is not necessary to fully retrain the entire model. CCVR tunes the classifier using virtual representations sampled from an approximate Gaussian mixture model. CReFF re-trains the classifier by utilizing a set of learnable federated features available on the server and optimizes the gradients of these federated features to align with the gradients of real data. FedCross utilizes multiple intermediate models for local training and adopts an effective cross-aggregation strategy to guide these intermediate models to search for a more generalized solution \cite{hu2024fedcross}.


Another commonly used approach is clustering-based client grouping, which aims to group clients into clusters based on their similarities in data or model characteristics to reduce the intra-cluster weight divergences. A straightforward method for clustered federated learning is to group models based on their loss values on local datasets \cite{ghosh2022efficient}. However, the significant computational cost hinders the deployment of this method. Some works measure the cosine similarity of local model parameters to form groups, but this approach may be inefficient in large-scale federated settings \cite{long2023multi}. Recent study has also developed a method to assess the differences in the feature space, which reduces computational costs while improving performance \cite{wei2023edge}.

\section{Preliminaries}\label{}
Suppose there are $N$ parties $P=\{P_1,..,P_N\}$, the party $P_n$ holds a private dataset $D_n=\{(x_1^n, y_1^n),...,(x_{N_n}^m, y_{N_n}^m)\}$ and a local model $F_n=E_n \odot H_n$, where $E_n$ and $H_n$ indicate the feature extractor and the classifier, respectively. $x_j^n$ is the j-th training sample of $D_n$, and $y_j^n \in \{1,2,...,K\}$ (K-classification task) is the label of $x_j^n$, $N_n$ represents the number of training samples in the dataset $D_n$. The server $S$ is utilized to coordinate collaborative training among multiple clients. Conventional methods typically involve three key stages: local training, cross-training, and model aggregation. They perform random broadcasting to exchange local models among parties, which can enlarge the training dataset of local models, i.e. execute  $D_n\stackrel{\text { training }}{\longrightarrow}F_n $ and $D_j\stackrel{\text { training }}{\longrightarrow}F_n$ sequentially.  


\begin{figure*}[t]
\centering

\includegraphics[width=1.0\textwidth]{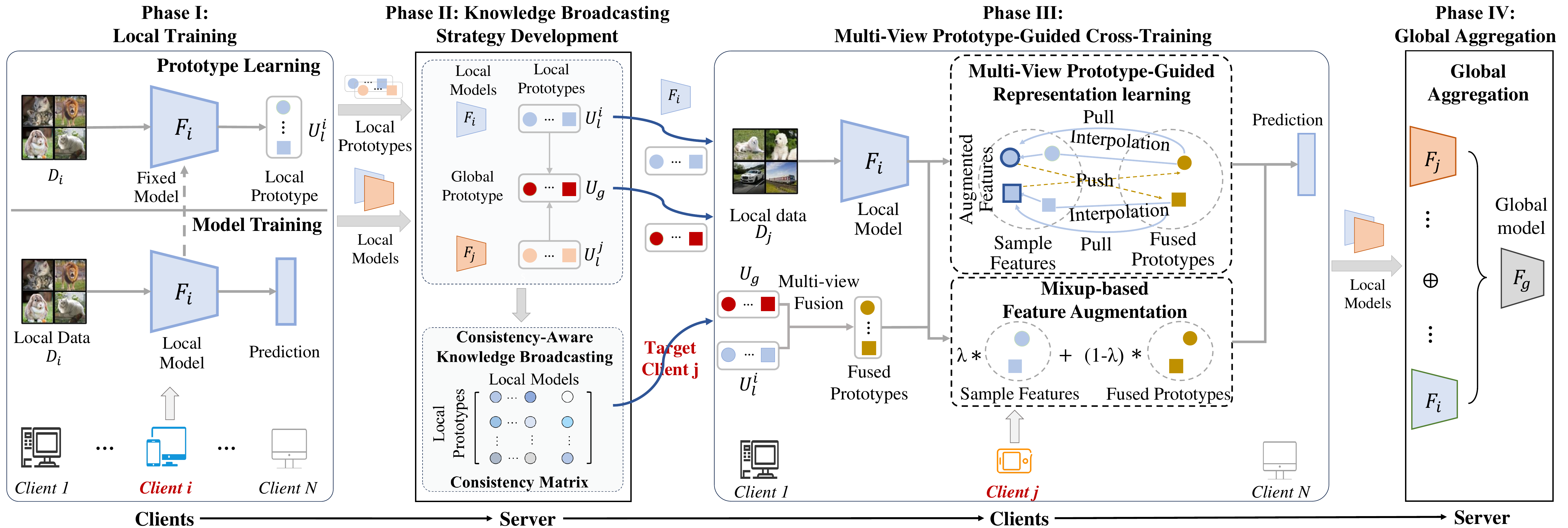}
\caption{Illustration of the FedCT framework, exemplified by cross training from client $i$ to client $j$. It can use arbitrary methods (such as FedAvg and MOON) to optimize local model in phase I, and generate a prototypical representation for each class. FedCT develops a knowledge broadcasting strategy in phase II, which utilizes the consistency of knowledge between clients to determine the path of knowledge broadcast. In phase III, FedCT leverages multi-view knowledge to guide the cross-training process, simultaneously alleviating the problem of knowledge forgetting from both local and global perspectives. Finally, FedCT aggregates all local models in phase IV. Notably, FedCT can perform multiple rounds of knowledge broadcasting and cross-training before global aggregation, i.e. exchange iterations $N_e$ can be modified.}
\label{fig3}
\end{figure*}

\section{Method}
\subsection{Overall Framework}

To mitigate the challenge of feature space heterogeneity introduced by data distribution discrepancies during cross-training, this paper proposes a multi-view knowledge-guided cross-training framework (FedCT). It leverages the personalized view knowledge to preserve client-specific semantics and enrich local knowledge, while incorporating the global view knowledge to provide consistent semantic anchors that facilitate representation alignment across clients. By integrating both views, the proposed mechanism enlarges the effective training sample set and facilitates consistent representation learning across clients under heterogeneous federated settings. Specifically, FedCT consists of four key phases: local training, consistency-aware knowledge broadcasting, multi-view guided cross training, and global aggregation, as shown in Figure \ref{fig3}.

\subsection{Phase I: Local Training}


FedCT optimizes local models and provides the prototypical information to the server without privacy leakage in the Local Training phase. It utilizes local data to train a local model and outputs a set of local prototypes. Notably, it can employ arbitrary client-based method to achieve this goal. Specifically, to assist in developing the local models assignment strategy, FedCT leverages pre-trained local models obtained by local training to extract local features and generate a class-aware prototype in representation space for each class.  For instance, the procedure to compute prototypes $u_i^k$ of class $k$ in the client $i$ can be expressed as follows:
\begin{equation}
u_i^k  = \textbf{mean}(E_i(x)|x \in D_i^k)
\end{equation}
where $E_i(\cdot)$ is an image encoder of client $i$, $D_i^k$ denote the data of class $k$ in client $i$.
Then all clients send their local prototype sets $U_l^i=\{u_i^k|k \in \{1,...,K\}\}$ to the server.

\subsection{Phase II: Knowledge Broadcasting Strategy Development}
FedCT designs a information broadcasting strategy in Knowledge Broadcasting phase, which guides the server to broadcast local models, local prototypes, and global prototypes to all clients based on predefined rules. The server obtains local models $\{F_n=E_n\odot H_n|n=1,...,N\}$ and prototype sets $\{U_l^n|n=1,...,N\}$ from all parties. To  foster the creation of consistent knowledge repositories between clients, the server aggregates all local prototypes to generate global prototypes,

\begin{equation}
    u_g^k =\frac{1}{N} \sum_{n=1}^{N} u_n^k
\end{equation}
where the global prototypes set can be denoted as $U_g=\{u_g^k|k=1,...,K\}$.

\vspace{0.1cm}
\subsubsection{Consistency-Aware Knowledge Broadcasting (CAKB).}
To promote effective collaboration under data heterogeneity, the CAKB module assigns each client a suitable partner for cross training, based on their knowledge compatibility. Since direct data sharing is not allowed, it is crucial to design a strategy that estimates the semantic similarity between clients without violating privacy. To achieve this, CAKB evaluates the consistency of knowledge among clients and outputs an assignment sequence for local models, local prototypes, and global prototypes. Intuitively, the consistency of knowledge can be measured by evaluating classification loss; however, private client data cannot be exposed to the server. To overcome this issue, we make class predictions for local prototypes using the classifiers of other clients, and compute the loss to estimate cross-client knowledge consistency, i.e.,



\begin{equation}
{Matrix}_{consist}  = \left[ {\begin{array}{*{20}c}
   {v_{11} } & {...} & {v_{1N} }  \\
   {...} & {...} & {...}  \\
   {v_{N1} } & {...} & {v_{NN} }  \\
\end{array}} \right]
\end{equation}
where $v_{ij}$ indicates the classification loss of classifier $H_i(\cdot)$ of client $i$ on the prototype $U_l^j$, i.e. $v_{ij}=\mathcal{CE}(H_i(U_l^j), Y_l^j)$, $Y_l^j$ denotes label set of prototype set $U_l^j$.

Based on the knowledge consistency learned by clients, we develops two rules, i.e.

\noindent
\textbf{Consistency Principle:} we assign the local model $F_j=E_j \odot H_j$, the local prototype $U_l^j$ and the global prototype $U_g$ to the target party $P_{target}^j$ with the highest knowledge consistency, which promotes cooperation between clients.
\begin{equation}
    P_{target}^j=\mathop {argmin }\limits_{\tilde P \in P\backslash P_j } [\mathop {\min }\limits_{i \in \tilde P} \mathcal{CE}(H_j(U_l^i),Y_l^i)]
\end{equation}

\noindent
\textbf{Inconsistency Principle:} we assign the local model, the local prototype and the global prototype to the target party $P_{target}^j$ with the lowest knowledge consistency, which may contribute to the expansion of the knowledge base.
\begin{equation}
    P_{target}^j=\mathop {argmax }\limits_{\tilde P \in P\backslash P_j } [\mathop {\max }\limits_{i \in \tilde P} \mathcal{CE}( H_j(U_l^i),Y_l^i)]
\end{equation}
And the target clients sequence can be denoted as $P_{target}=\{P_{target}^j|j=1,...,N\}$. Subsequently, the server broadcasts the local model $F_j$, local prototype $U_l^j$, and global prototype $U_g$ to the target client $P_{target}^j$ based on the proposed principles.

\subsection{Phase III: Multi-View Knowledge-Guided Cross Training}
FedCT retrains local models under the guidance of multi-view prototypical knowledge in Cross Train phase, which helps the local model learn comprehensive knowledge. In this phase, each client obtains the local model, local prototype and global prototype from other client based on consistency-aware knowledge broadcast rules. It contains two main modules in the training process: the multi-view knowledge-guided representation learning module  and the mixup-based feature augmentation module. 

\subsubsection{Multi-View Knowledge-Guided Representation Learning (MVKGRL).}
The MVKGRL module aims to address the inconsistency in feature representations caused by data heterogeneity across clients. To this end, it introduces multi-view prototypes as guidance signals to encourage clients to learn semantically aligned class-level representations, where the local view captures client-specific semantics to preserve personalized knowledge, and the global view provides aggregated semantic anchors that promote cross-client consistency.
Specifically, it first fuses personalized and global prototypes, and then extends traditional contrastive learning, and leverages augmented samples to enhance the effectiveness of representation learning, which aligns the distribution of samples representations and the corresponding prototypes.



As illustrated in Figure \ref{fig3}, after the knowledge broadcasting phase, the MVKGRL module in client $j$ acquires the local model $F_i$, local prototypes $U_l^i$ and global prototypes $U_g$ from the server. To ensure distinctiveness of representations in the latent space between classes, FedCT encourages the learning of unified features by maximizing the agreement between each sample and its corresponding prototype. The multi-view prototypical fusion can be represented as:

\begin{equation}
    U_f = \lambda_{fuse} \times  U_g + (1-\lambda_{fuse}) \times U_l^i
\end{equation}
where $U_f=\{u_f^k|k=1,...,K\}$ is the fused prototype, $\lambda_{fuse}$ is a weighted parameter.

Inspired by contrastive learning in representation learning, we define the augmented prototypical contrastive loss:
\begin{small}
\begin{equation}
\begin{split}
    &\mathcal{L}_{APCL}  = \\ & - \log \frac{{\exp (sim(f_{hybrid},u_f^ +  )/\tau_2 )}}{{\exp (sim(f_{hybrid},u_f^ +  )/\tau_2 ) + \sum {\exp (sim(f_{hybrid},u_f^ -  )/\tau_2 )} }}
\end{split}
\end{equation}    
\end{small}


where $f_{hybrid}=\lambda_{hy} \times (f_x-u_f^+) + f_x$, $u_f^+$ and $u_f^-$ denote the prototype with the same and different labels as feature $f_x$ respectively. $sim(\cdot)$ is cosine similarity function, $\tau_2$ is a temperature parameter. It is worth highlighting that the use of extrapolation can effectively promote the alignment between the features of hard samples and prototypes.

\subsubsection{Mixup-based Feature Augmentation (MFA).}
The MFA module aims to exploit the historical features to improve the generalization of local models. A practical idea is to reuse the class-aware prototypes to refine and augment image features. Specifically, it uses mixup to generate augmented features $f_{mix}$, it has the label $y_{mix}$, defined by 
\begin{equation}
f_{mix}=\lambda_{mix}  \times f_a + (1-\lambda_{mix})  \times f_b
\end{equation}
\begin{equation}
y_{mix}=\lambda_{mix} \times y_a + (1-\lambda_{mix})  \times y_b
\end{equation}
where $f_a$ and $f_b$ are the representations of two different samples, $y_a$ and $y_b$ are their labels. $\lambda_{mix}$ is a constant coefficient. And we use an mixup loss to optimize the model,
\begin{equation}
    \mathcal{L}_{mix} = \lambda_{mix} \times CE(\tilde {y}_a, y_a) + (1-\lambda_{mix}) \times CE(\tilde {y}_b, y_b)
\end{equation}
where $\tilde{y}_a$ and $\tilde{y}_b$ represent the predictions of corresponding samples.

\begin{table*}[t]
\centering
\caption{Performance comparison between FedCT with baselines on CIFAR10, CIFAR100, TinyImagenet, and VireoFood172 datasets. All methods were conducted over three trials, with reports including both the average and the standard deviation.}
\label{tab2}
\setlength{\tabcolsep}{3mm}{
\begin{tabular}{c|c|c|c|c|c|c|c|c}
\hline \multirow{2}{*}{ Methods } & \multicolumn{2}{c|}{ CIFAR10 } & \multicolumn{2}{c|}{ CIFAR100 } & \multicolumn{2}{c|}{ TinyImagenet } & \multicolumn{2}{c}{ VireoFod172 } \\
\cline { 2 - 9 } &  $\beta$=0.1  &  $\beta$=0.5  &  $\beta$=0.1  &  $\beta$=0.5  &  $\beta$=0.1  &  $\beta$=0.5  &  $\beta$=0.1  &  $\beta$=0.5  \\
\hline 
FedAvg & 74.37±0.6 & 84.12±0.7 & 53.16±0.6 & 53.58±0.9 & 44.46±0.4 & 47.72±0.7 & 61.32±1.1 & 66.56±0.9 \\
MOON & 75.79±0.8 & 85.21±0.4 & 53.54±0.9 & 55.82±0.5 & 45.04±0.8 & 48.52±0.6 & 62.28±0.7 & 67.18±0.8 \\
FedNTD & 75.42±1.3 & 84.68±0.6 & 53.81±0.8 & 55.27±1.1 & 46.08±0.5 & 48.19±0.9 & 62.22±0.8 & 67.42±0.5 \\
FedProc & 75.08±0.7 & 85.49±0.6 & 54.03±0.9 & 56.02±0.7 & 45.51±0.8 & 47.69±0.5 & 61.46±0.3 & 66.88±0.8 \\
FedDecorr & 75.64±1.0 & 85.37±0.6 & 54.12±0.8 & 55.74±0.9 & 45.84±0.8 & 48.07±0.7 & 62.59±0.6 & 67.63±0.5 \\
CLIP2FL & 76.78±0.9 & 86.39±0.5 & 55.62±0.8 & 56.27±0.7 & 47.16±0.6 & 47.94±0.8 & 61.65±0.4 & 67.83±0.9 \\
FedRCL & 75.11±0.8 & 84.43±0.4 & 54.12±0.7 & 54.45±1.1 & 45.31±0.7 & 48.93±0.6 & 61.68±0.5 & 67.23±0.7 \\
FedFSA & 77.68±0.5 & 86.03±0.8 & 55.39±0.9 & 56.93±1.0 & 45.85±0.6 & 45.81±0.5 & 61.68±0.6 & 67.84±0.8 \\
FedDDL & 77.41±0.7 & 87.21±1.1 & 56.13±0.7 & 57.23±1.4 & 44.27±0.8 & 44.49±0.8 & 61.12±0.7 & 67.43±0.6 \\
FedExg & 74.88±0.8 & 86.04±0.7 & 54.93±1.1 & 55.43±0.7 & 46.13±0.8 & 49.26±0.7 & 62.15±0.6 & 67.68±0.4 \\
PGCT & 76.23±0.8 & 86.62±0.5 & 55.74±0.6 & 56.12±0.9 & 47.11±0.3 & 50.79±0.4 & 62.42±0.8 & 68.04±0.3 \\
\hline 
FedCT$_{\rm{FedAvg}}$ & 78.22±0.5 & 88.21±0.4 & 57.07±0.7 & 58.23±0.6 & \textbf{48.63±0.3} & 52.76±0.7 & 63.73±0.8 & 69.34±0.4 \\
FedCT$_{\rm{MOON}}$ &  \textbf{78.68±0.4}  &  \textbf{88.91±0.3}  &  \textbf{57.43±0.8}  &  \textbf{58.94±0.5}  &  48.51±0.6  &  \textbf{53.29±0.9}  &  \textbf{63.84±0.5}  &  \textbf{69.61±0.3}  \\
\hline
\end{tabular}}
\end{table*}

\subsection{Phase IV: Global Aggregation}
FedCT aggregates all local models uploaded by clients to build the global model in the Global Aggregation phase. The commonly used method is weighted averaging, i.e.,
\begin{equation}
   w_{G}^{t+1} \leftarrow \sum_{n=1}^{N} \frac{\left|D_{n}\right|}{|D|} w_{n}^{t}
\end{equation}
where $w_{G}^{t+1}$ denotes the parameter of the initial (global) model of round $t+1$, and $w_{n}^{t}$ is the parameter of $n$-th local model in round $t$. $\left|D_{n}\right|$ is the size of local dataset $D_n$ and $|D|=\sum_{n=1}^{N}\left|D_{n}\right|$. Notably, as a model-agnostic approach, FedCT can combine various methods, such as FedNova and FedEA, to perform the aggregation process.

\subsection{Training Strategies}
FedCT focuses on guiding local models to learn comprehensive knowledge. It has two independent training phases, including phases I and III. Its training strategy is as follows:
\begin{itemize}[leftmargin=10pt]
    \item \textbf{In phase I}, FedCT needs to determine its optimization objectives based on the methods it employs,
    \begin{equation}
        \mathcal{L}_{phase1}= \mathcal{L}_{base}
    \end{equation}
    where the base could be FedAvg, MOON, and so on.
    
    \item \textbf{In phase III}, FedCT aims to guide all clients to learn homogeneous representations, and its objective is to minimize the following loss: 
    \begin{equation}
        \mathcal{L}_{phase3} = \mathcal{L}_{cls} + \kappa \times\mathcal{L}_{APCL} + \eta \times \mathcal{L}_{mix}
    \end{equation}
    where $\kappa$ and $\gamma$ are weight parameters.

\end{itemize}



\section{Experiments}
\subsection{Experiment Settings}
\subsubsection{Datasets.}
To verify the effectiveness of the proposed FedCT, extensive experiments were conducted  on three commonly used datasets in federated learning (CIFAR10 \cite{krizhevsky2009learning}, CIFAR100 \cite{krizhevsky2009learning}, TinyImagenet \cite{le2015tiny}) and one challenging food classification dataset (VireoFood172 \cite{chen2016deep}). 

\subsubsection{Network Architecture.}
To ensure fair comparison, all methods use ResNet-18 \cite{he2016deep} as the backbone. Following prior work \cite{gao2022feddc,li2021model}, we reduce the first convolutional kernel size from 7 to 3 for CIFAR-10, CIFAR-100, and TinyImageNet, while keeping it at 7 for VireoFood172.

\subsubsection{Hyper-parameter Settings.}
In all methods, we conducted the local training for $10$ epochs in each global round, with a total of $10$ clients and a sample fraction of $1.0$. The local optimizer used was the SGD algorithm, and the communication round was set to $100$. During local training and cross training phases, we set the weight decay to $1e-05$ and used a batch size of $64$. The learning rate was initialized to $0.01$, and the Dirichlet parameter $\beta$ was set to $0.1$ and $0.5$, while the temperature parameter $\tau_1$ and $\tau_2$ were selected from $\{0.5,0.05\}$ and $\eta \in \{0.1,0.5\}$, $\kappa \in \{0.5,1.0,2.0,5.0\}$, $\gamma \in \{0.01, 0.05, 0.1,0.5\}$, and $\lambda_{hy}$ and $\lambda_{mix}$ were both tuned from $\{0.1,0.3,0.5\}$, $\lambda_{fuse}$ is set to 0.5. And the exchange iterations $N_e$ is set 1. The settings for other hyper-parameters were based on the corresponding paper.

\subsection{Performance Comparison}
We conduct comparisons between FedCT and existing methods, categorized into three groups: 1) federated learning (FL) methods without cross-training, including FedAvg \cite{mcmahan2017communication}, FedProx \cite{li2020federated}, MOON \cite{li2021model}, FedProc \cite{mu2023fedproc}, FedDecorr \cite{shi2022towards}, CLIP2FL \cite{shi2024clip}, FedRCL \cite{seo2024relaxed}; FedFSA \cite{qi2025cross}; FedDDL \cite{qi2025federated}; 2) FL methods with cross-training, including FedExg \cite{mao2020fedexg} and PGCT \cite{liucross}. The following results presented in Table \ref{tab2}: 
\begin{itemize}[leftmargin=10pt]
    \item FedCT$_\text{FedAvg}$ and FedCT$_\text{MOON}$ show remarkable improvements in federated classification performance compared to the corresponding baselines, which highlights the model-agnostic character of the proposed FedCT.

    \item Combining cross-training typically leads to superior performance compared to baseline methods. This is understandable since cross-training expands the learnable knowledge repository of the local model.
    
    \item The proposed FedCT usually outperforms other cross-training methods. This is reasonable because FedCT not only enlarges the training set of the local model but also incentivizes clients to construct a unified knowledge repository, which highlights the greater importance of mitigating global knowledge forgetting over local forgetting.
    
    
    \item FedCT maintains superior performance in cases with different levels of heterogeneity, which indicates that learning multi-view knowledge from the representation space is a viable approach.
                         
\end{itemize}


\subsection{Ablation Study}
This section further investigates the effectiveness of the key modules of FedCT on the CIFAR10 and CIFAR100 datasets with different levels heterogeneity $\beta=0.1$ and $\beta=0.5$. The results are extensively detailed in Table \ref{tab3} and \ref{tab4}, which highlights the performance outcomes on various cases.

\begin{table}[h]
\centering
\caption{Ablation study of FedCT on CIFAR10 and CIFAR100 datasets with different heterogeneity $\beta=0.1$ and $\beta=0.5$.}
\label{tab3}
\scalebox{0.8}{
\begin{tabular}{c|c|c|c|c}
\hline
\multirow{2}{*}{} & \multicolumn{2}{c|}{\textbf{CIFAR10}} & \multicolumn{2}{c}{\textbf{CIFAR100}} \\ \hline
                  & $\beta=0.1$          & $\beta=0.5$          & $\beta=0.1$           & $\beta=0.5$          \\ \hline
Base            & 74.37±0.6             & 84.12±0.8             & 53.16±0.5              & 53.58±0.7             \\
+CAKB              & 75.42±0.4             & 86.62±0.6             & 55.17±0.5              & 55.81±0.7             \\
+CAKB+MFA         &  76.02±0.6            & 86.89±0.4             & 55.68±0.6              & 56.47±0.5             \\
+CAKB+MVKGRL        & 77.36±0.7             & 87.42±0.4             & 56.37±0.6              & 57.19±0.5             \\
+CAKB+MVKGRL+MFA & \textbf{78.22±0.5}             & \textbf{88.21±0.4}            & \textbf{57.07±0.7}              & \textbf{58.33±0.6}        \\ \hline     
\end{tabular}}
\end{table}

\begin{table}[t]
\centering
\caption{The influence of knowledge broadcasting strategies on the final performance is examined on the CIFAR-10 and CIFAR-100 datasets with $\beta=0.5$. Furthermore, the communication rounds required to achieve an accuracy of 88.0 on CIFAR-10 and 57.0 on CIFAR-100 are reported.}
\label{tab4}
\begin{tabular}{c|c|c|c|c}
\hline
\multirow{2}{*}{} & \multicolumn{2}{c|}{CIFAR10} & \multicolumn{2}{c}{CIFAR100} \\ \hline
                  & Acc          & Rounds       & Acc          & Rounds        \\ \hline
Random            & 87.78±0.7        & 104           & 57.68±0.4        & 76            \\
CAKB$_{Consistency}$               & \textbf{88.21±0.4}        & \textbf{77}           & \textbf{58.33±0.6}        & \textbf{48}            \\
CAKB$_{Inconsistency}$             & 87.79±0.7        & 122          & 57.53±0.6        & 82    \\\hline        
\end{tabular}
\end{table}

\begin{itemize}[leftmargin=10pt]
    \item Exclusively depending on random knowledge broadcasting (Exg) might not yield a substantial improvement. This is due to the heterogeneity of data distributions between clients, which leads to the local models being prone to forgetting global knowledge during multiple rounds of cross-training. 
    
    \item Cross training, when combined with the mixup-based feature augmentation (MFA) module, leads to performance improvement. The MFA module leverages prototype information from mixed features to implicitly regulate the construction of consistent knowledge between clients and enhance the discriminative capability of classifier.
   
    \item Benefiting from the guidance of the multi-view knowledge-guided representation learning (MVKGRL) module, the combination of model exchange and the MVKGRL module outperforms baseline method on both datasets, which achieves a substantial margin of 2.5\% and 3.6\%. 
    

    \item As depicted in Table \ref{tab4}, inconsistency-based broadcasting results in poor performance. This could be due to the disparities among the knowledge, which hinder the model from fitting data from multiple sources simultaneously.

    \item Consistency-based broadcasting can accelerate the convergence of the model. This also reflects the importance of acquiring consistent knowledge in fostering collaboration among clients.
    
\end{itemize}

\subsection{In-depth Analysis}
\begin{figure}[h]
\centering
\includegraphics[width=0.48\textwidth]{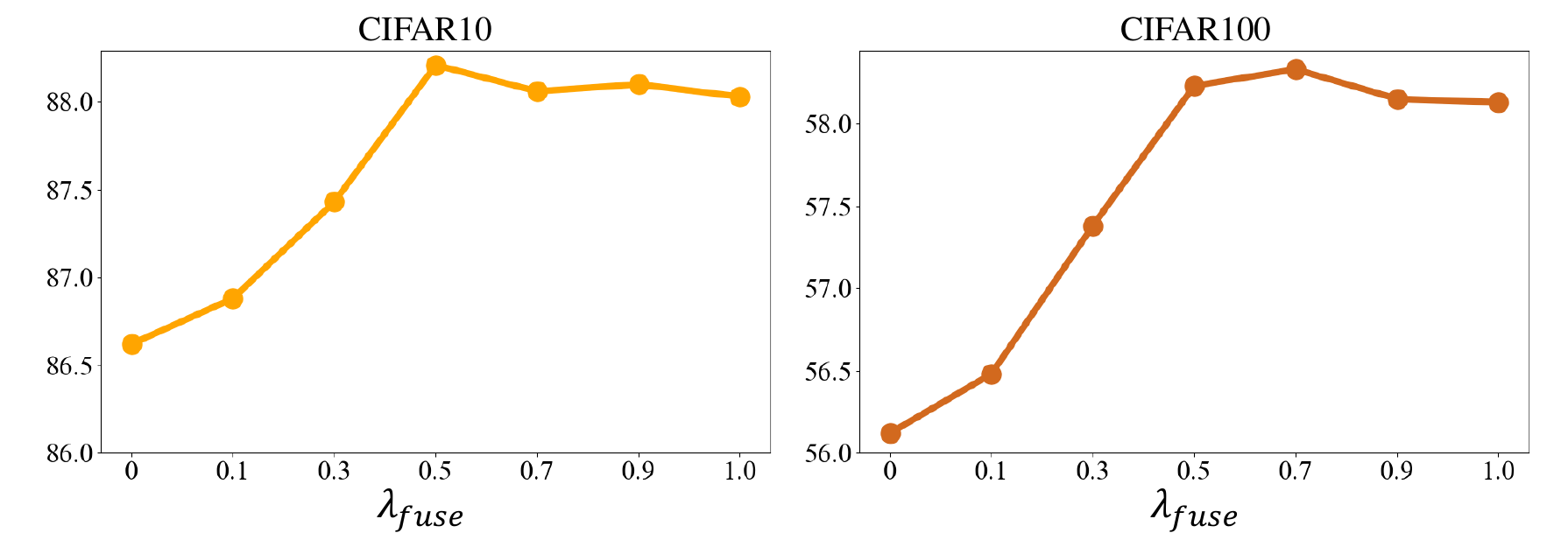}
\caption{The impact of applying different-view knowledge on performance ($\lambda_{fuse}$=$\{0,0.1,0.3,0.5,0.7,0.9,1.0\}$) on the CIFAR10 and CIFAR100 datasets with the heterogeneity $\beta=0.5$.} 
\label{fig4}
\end{figure}
\subsubsection{The impact of applying the multi-view knowledge on performance.} 


This section evaluates the impact of applying different-view knowledge by varying the fusion weight $\lambda_{fuse}$ in $\{0, 0.1, 0.3, 0.5, 0.7, 0.9, 1.0\}$. $\lambda_{fuse}=0$ and $\lambda_{fuse}=1.0$ correspond to using only local-view and only global-view knowledge, respectively. Experiments are conducted on CIFAR10 and CIFAR100 under heterogeneity $\beta=0.5$. As shown in Figure~\ref{fig4}, multi-view fused knowledge consistently outperforms local-view only, as it enables local models to incorporate richer global information and promotes a consistent knowledge base across clients. The results suggest that global knowledge is more critical than local knowledge in the MVKGRL module, indicating that global forgetting is a key bottleneck in federated learning. The best performance is observed at $\lambda_{fuse}=0.5$ on CIFAR10 and $\lambda_{fuse}=0.7$ on CIFAR100, suggesting an effective balance between local and global views. 

\subsubsection{The influence of the number of exchange iterations on performance.}

\begin{figure}[t]
\centering
\includegraphics[width=0.48\textwidth]{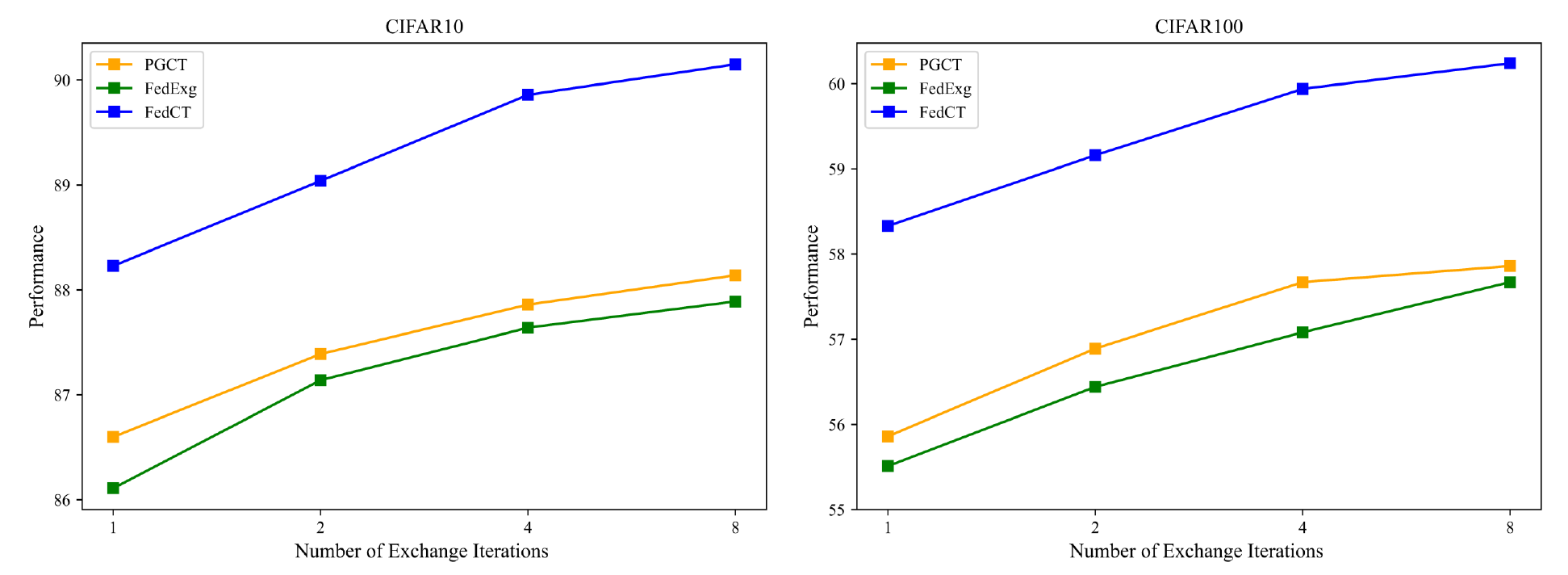}
\caption{The influence of the number of exchanege iterations
($N_e$=1,2,4,8) on the final performance of cross-training methods  on the CIFAR10 and CIFAR100 datasets with the heterogeneity $\beta=0.5$.} 
\label{fig5}
\end{figure}

We evaluate the effect of the exchange iteration number $N_e$ on model performance, using CIFAR10 and CIFAR100 datasets with $\beta=0.5$. $N_e$ is varied in $\{1, 2, 4, 8\}$, while the total number of communication rounds is fixed at 100.
As shown in Figure~\ref{fig5}, increasing the number of exchange iterations generally improves model performance, but the gains gradually plateau. This is expected, as performance depends not only on the exchange iterations but also on the aggregation frequency. Given a fixed number of communication rounds, more exchange iterations reduce aggregation frequency, which may limit further improvements.


\subsubsection{The effectiveness of FedCT on different levels heterogeneity.}

\begin{table}[h]
\centering
\caption{Performance comparison of various cross-training methods in scenarios with varying data heterogeneity.}
\label{tab6}
\setlength{\tabcolsep}{1.8mm}{
\begin{tabular}{c|c|c|c|c|c}
\hline
\multirow{2}{*}{Methods} & \multicolumn{5}{c}{\textbf{CIFAR10}}           \\\cline{2-6}
                         & $\beta=0.05$  & $\beta=0.1$   & $\beta=0.5$   & $\beta=1.0$   & $\beta=5.0$   \\ \hline
FedExg                   & 57.56±0.8 & 74.08±0.8 & 86.04±0.7 & 88.98±0.6 & 89.83±0.5 \\
PGCT                     & 58.03±0.8 & 75.93±0.8 & 86.62±0.5 & 89.43±0.6 & 90.16±0.4 \\
FedCT                    & \textbf{59.78±0.6} & \textbf{78.22±0.5} & \textbf{88.21±0.4} & \textbf{90.70±0.8} & \textbf{91.25±0.5} \\\hline
\multirow{2}{*}{Methods}        & \multicolumn{5}{c}{\textbf{CIFAR100}}           \\\cline{2-6}
                         & $\beta=0.05$  & $\beta=0.1$   & $\beta=0.5$   & $\beta=1.0$   & $\beta=5.0$   \\ \hline
FedExg                   & 51.83±1.4      & 54.93±1.1 & 55.21±0.7 & 55.46±0.6 & 55.78±0.8 \\
PGCT                     & 52.43±0.8      & 55.74±0.6 & 56.12±0.9 & 56.25±0.5 & 56.63±0.4 \\
FedCT                    & \textbf{52.87±0.7}      & \textbf{57.07±0.7} & \textbf{58.33±0.6} & \textbf{58.42±0.5} & \textbf{59.34±0.3} \\\hline
\end{tabular}}
\end{table}


This section evaluates the effectiveness of FedCT under different levels of heterogeneity by varying $\beta$ in $\{0.05, 0.1, 0.5, 1.0, 5.0\}$. Experiments are conducted on the CIFAR100 dataset. Table~\ref{tab6} shows that FedCT consistently outperforms existing cross-training methods across all settings, demonstrating its effectiveness under varying data heterogeneity. Notably, FedCT achieves the largest accuracy gain when $\beta=0.05$, indicating strong performance in highly heterogeneous scenarios. In addition, as $\beta$ increases, all methods exhibit improved accuracy.


\subsubsection{The scalability of FedCT.}

\begin{figure}[h]
\centering
\includegraphics[width=0.5\textwidth]{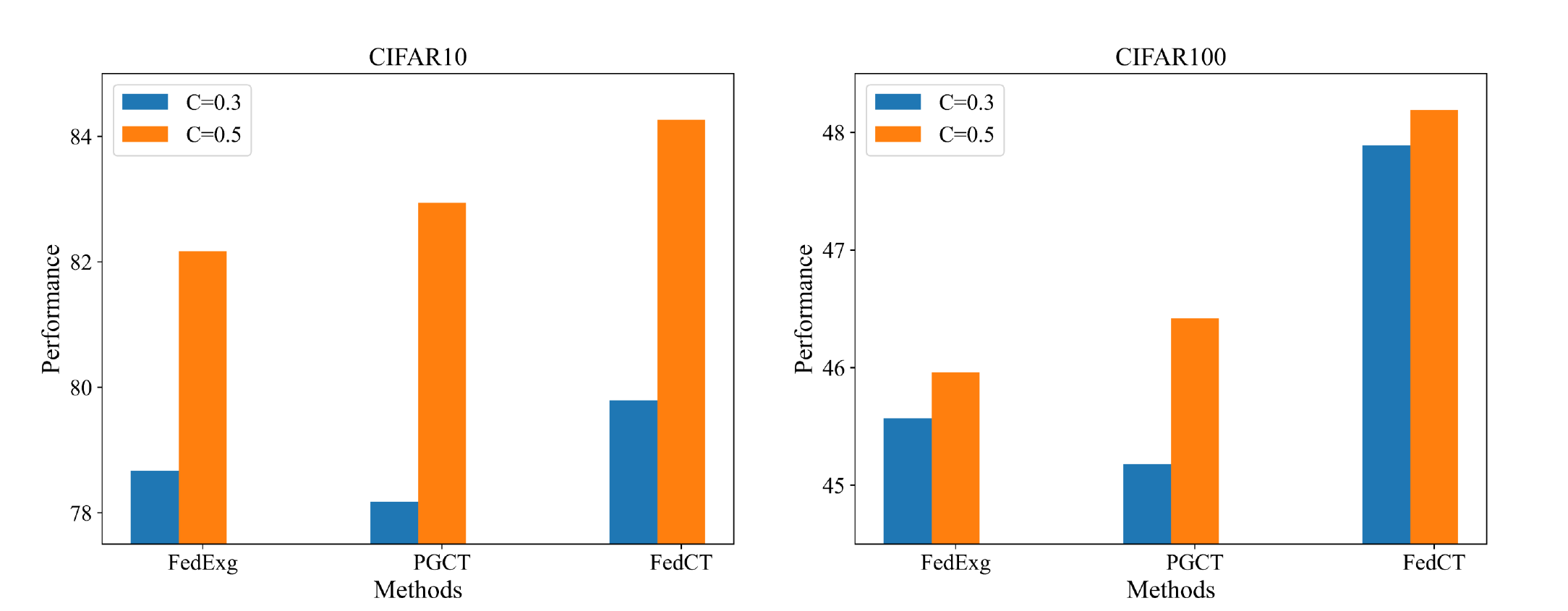}
\caption{The scalability of cross-training methods with partial client participant ($C=0.3$ and $C=0.5$). The number of clients was set to 100.} 
\label{fig6}
\end{figure}


This section investigates the scalability of FedCT by increasing the number of participants, with only a subset involved in each training round. Specifically, we set $N=100$, $\beta=0.5$, and 200 communication rounds, varying the client fraction $C$ in $\{0.3, 0.5\}$. Experiments are conducted on CIFAR10 and CIFAR100.
Figure~\ref{fig6} shows that FedCT consistently improves performance across different client participation rates by promoting consistent local representations. At $C=0.3$, it surpasses FedExg and PGCT by 1.2\% and 2\%, respectively; at $C=0.5$, the gains are 2\% and 1.4\%. These results underscore the importance of addressing global knowledge forgetting in federated learning.


\begin{figure}[h]
\centering
\includegraphics[width=0.48\textwidth]{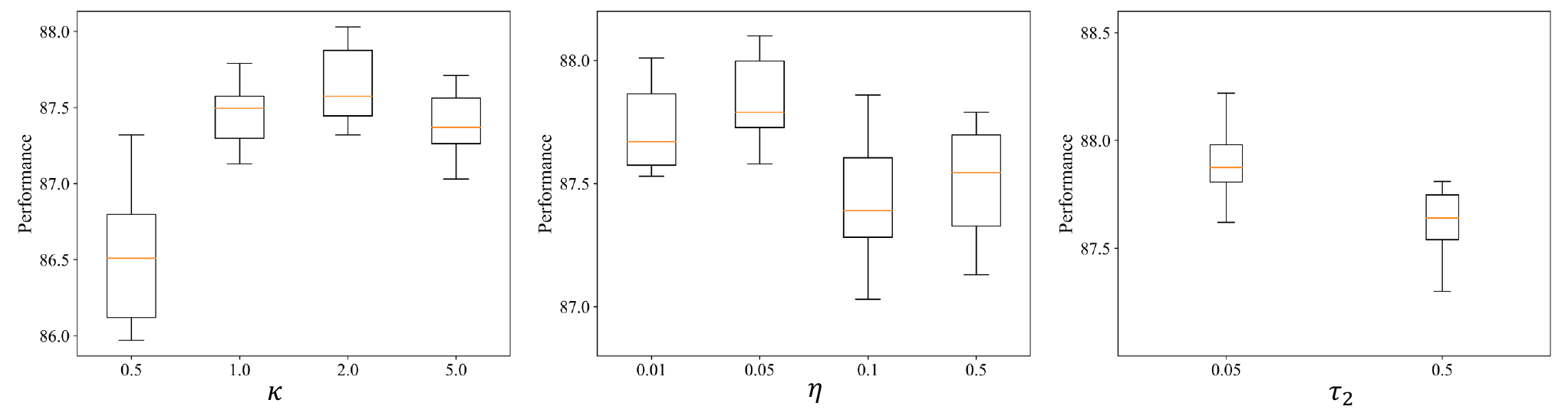}
\caption{The performance of FedCT using different hyperparameters $\kappa$, $\eta$ and $\tau_2$ on CIFAR10 with $\beta=0.5$.} 
\label{fig7}
\end{figure}

\subsubsection{Robustness of FedCT on hyperparameters.} 

This section evaluates the impact of hyperparameter choices. Specifically, $\kappa$ and $\eta$ are selected from $\{0.5, 1.0, 2.0, 5.0\}$ and $\{0.01, 0.05, 0.1, 0.5\}$, respectively, while the temperature parameter $\tau_2$ is tuned from $\{0.05, 0.5\}$. Figure~\ref{fig7} presents the test accuracy as a box plot. FedCT shows consistently stable performance across all settings, except for a slight drop when $\kappa=0.5$. This suggests that FedCT is robust and largely insensitive to hyperparameter choices within a broad range. Notably, even the worst cases still outperform existing cross-training methods.


\begin{table}[t]
\centering
\caption{The evaluation of different federated learning methods on CIFAR10 and CIFAR100 datasets ($\beta = 0.1$ and $\beta = 0.5$) in terms of the communication rounds needed to achieve the target test accuracy (Acc).}
\label{tab7}
\scalebox{0.8}{
\begin{tabular}{c|c|c|c|c}
\hline
              & \multicolumn{2}{c|}{\textbf{CIFAR10}}                                                                                     & \multicolumn{2}{c}{\textbf{CIFAR100}}                                                                                    \\ \hline
              & \begin{tabular}[c]{@{}c@{}}$\beta=0.1$\\ Acc=76.0\end{tabular} & \begin{tabular}[c]{@{}c@{}}$\beta=0.5$\\ Acc=85.0\end{tabular} & \begin{tabular}[c]{@{}c@{}}$\beta=0.1$\\ Acc=54.0\end{tabular} & \begin{tabular}[c]{@{}c@{}}$\beta=0.5$\\ Acc=55.0\end{tabular} \\ \hline
FedAvg        & 127                                                    & 107                                                    & 141                                                    & 127                                                    \\
MOON          & 106                                                    & 74                                                     & 122                                                     & 77                                                     \\
FedExg        & 66                                                     & 58                                                     & 76                                                     & 64                                                     \\
PGCT          & 56                                                     & 52                                                     & 52                                                     & 62                                                     \\
FedCT(FedAvg) & \textbf{28}                                                     & \textbf{24}                                                     & \textbf{36}                                                     & \textbf{26}                                                     \\
FedCT(MOON)   & 34                                                     & 38                                                     & 46                                                     & 34                                                    \\ \hline
\end{tabular}}
\end{table}

\subsubsection{Communication Efficiency.}
This section examines the number of communication rounds needed by different methods to reach the target test accuracy. As shown in Table~\ref{tab7}, FedCT achieves the best results on both datasets, reducing the required rounds compared to all baselines. This highlights that mitigating global knowledge forgetting improves federated model performance. Notably, FedCT$_{\text{FedAvg}}$ achieves an acceleration ratio of nearly 4× on CIFAR10, demonstrating significantly higher communication efficiency than existing methods.

\subsection{Case Study}
\begin{figure}[h]
\centering
\includegraphics[width=0.5\textwidth]{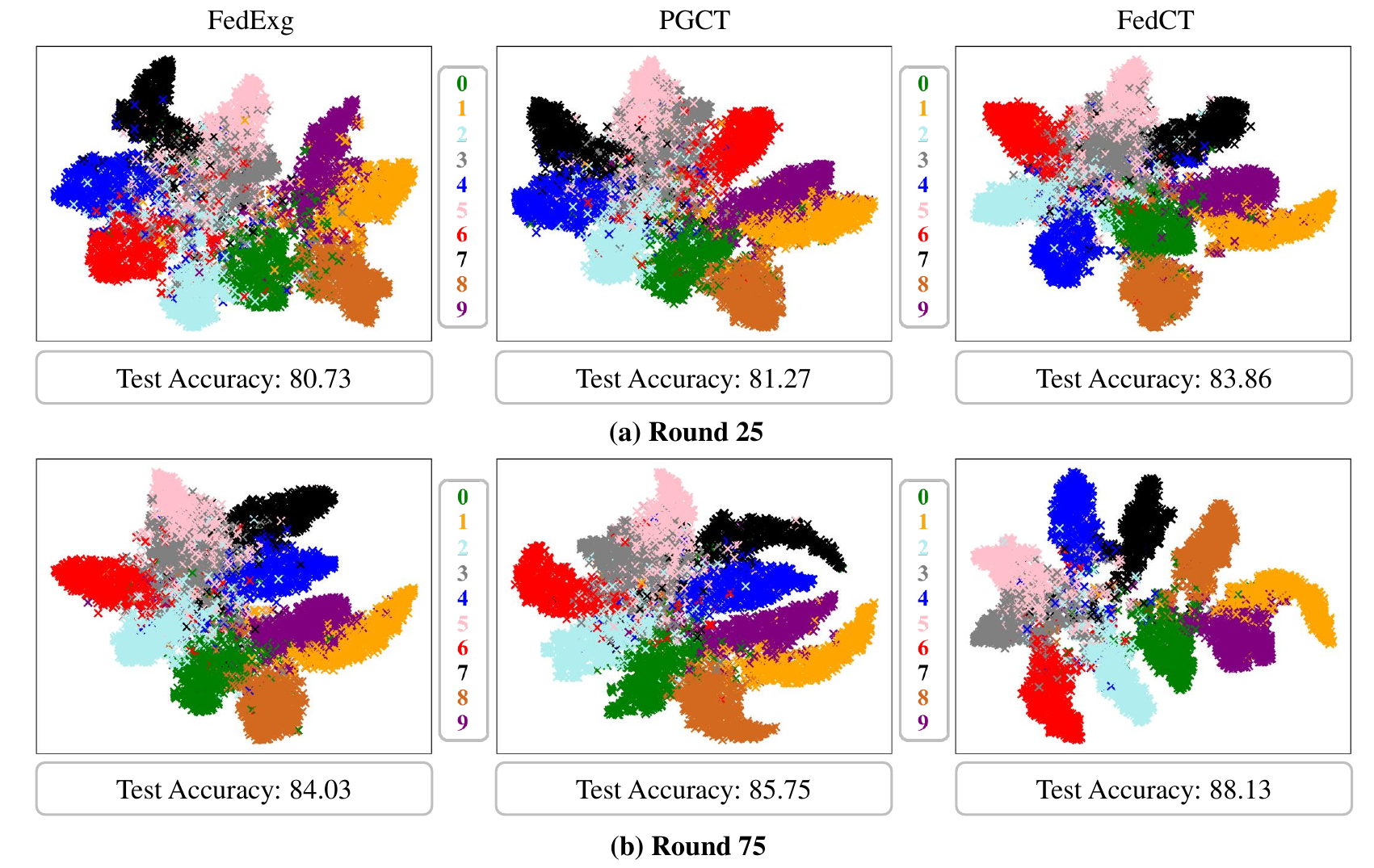}
\caption{Visualization of the representations and performance outputs  at different rounds varies across different methods. FedCT learns the best data distribution and achieves optimal performance.} 
\label{fig8}
\end{figure}

\begin{figure*}[t]
\centering
\includegraphics[width=1.0\textwidth]{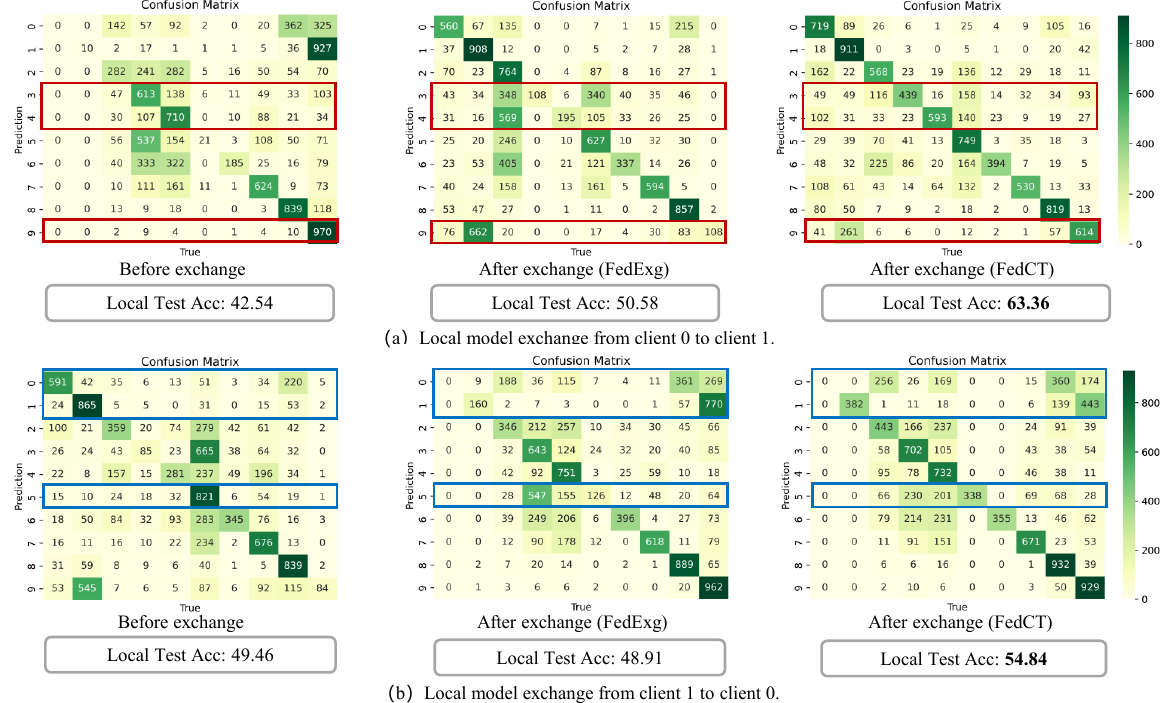}
\caption{Quantitative analysis of knowledge preservation. (a) Exchange the local model from client 0 to client 1. (b) Exchange the local model from client 1 to client 0. The proposed FedCT method alleviates the issue of knowledge forgetting in cross-training and improves its generalization.} 
\label{fig9}
\end{figure*}


\subsubsection{Quality Analysis of Representation Learning.}
This section investigates the representation space of FedCT and baseline methods. As shown in Figure~\ref{fig8}, we use t-SNE \cite{van2008visualizing} to visualize latent representations on the CIFAR10 test set, using results randomly selected from two training rounds. FedCT clearly outperforms FedExg and PGCT, as it yields more distinct decision boundaries between class representations. In contrast, FedExg exhibits significant feature overlap across classes, reflecting poor representation learning due to severe knowledge forgetting during training. PGCT alleviates this issue by preserving local knowledge, thereby improving generalization and representation quality, which aligns with expectations. However, PGCT only enforces intra-model consistency and overlooks client-specific data variability. FedCT addresses both local and global knowledge forgetting by fully leveraging multi-view information. These results confirm that the Multi-View Knowledge-Guided Representation Learning module enhances the model’s generalization ability.

\subsubsection{Quantitative Analysis of Knowledge Preservation.} 
This section presents a quantitative analysis of knowledge preservation before and after local model exchange across different methods. Two clients are randomly selected, with their data distributions shown at the top of Figure~\ref{fig9}. Each client first performs local training, followed by cross-training using the exchanged models. Overall, FedCT effectively mitigates knowledge forgetting during cross-training, preserving the classification ability of local models after exchange. As shown in Figure~\ref{fig9}, FedExg struggles with classes 3, 4, and 9 (red box) and 1 and 5 (blue box) due to limited samples, while FedCT improves recognition by leveraging prototypical knowledge. Both methods underperform on class 0, highlighting that data scarcity limits performance even with strong guidance—emphasizing the importance of data quantity. Overall, FedCT improves test accuracy by 12.78\% and 5.93\% over FedExg, demonstrating its superior generalization through prototype-guided cross-training.

\begin{figure*}[t]
\centering
\includegraphics[width=1.0\textwidth]{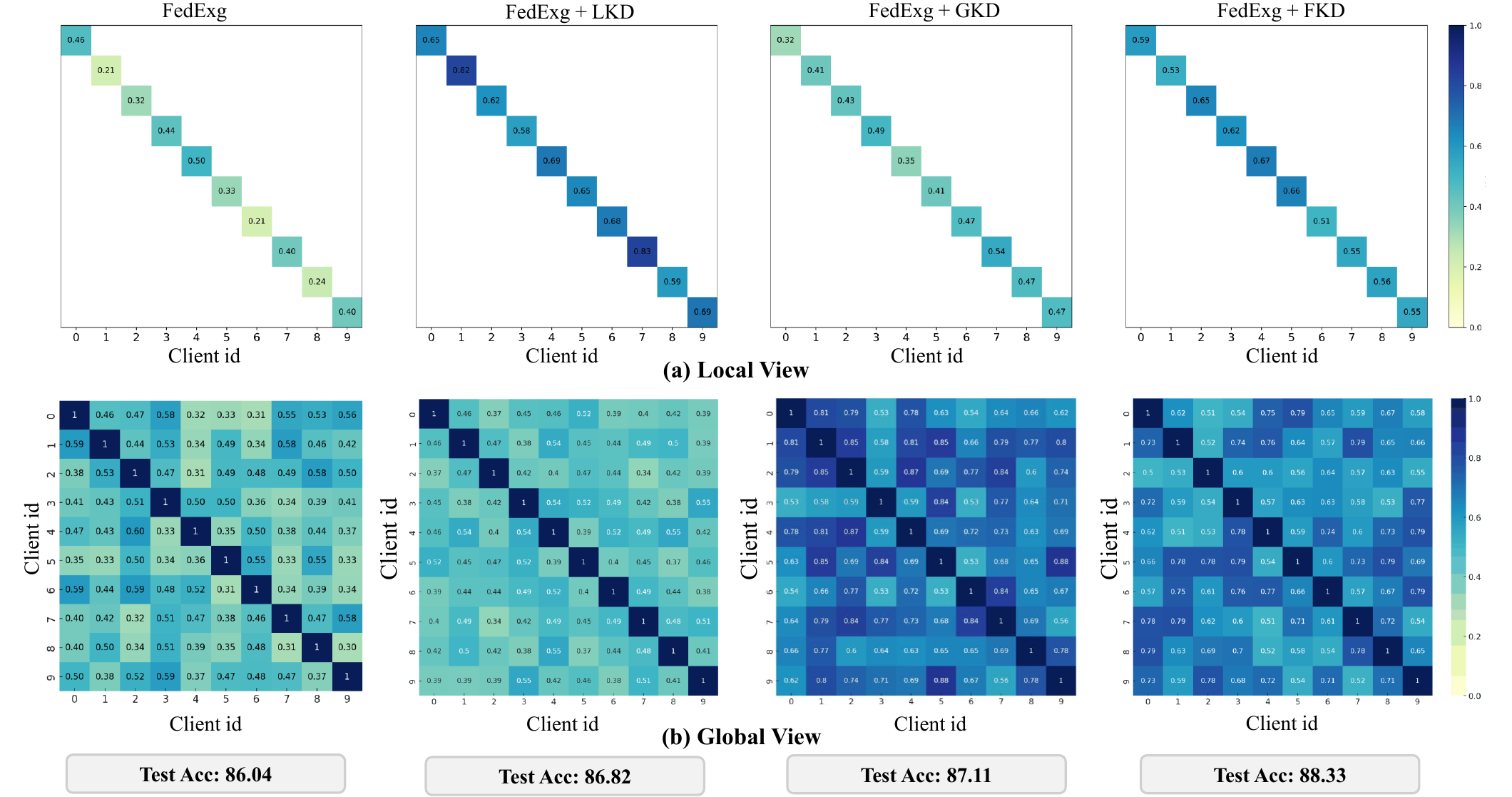}
\caption{Qualitative analysis of knowledge preservation. (a) Analyzing the extent of knowledge preservation  from a local view involves evaluating the CKA similarity between corresponding layers of the local models before and after exchange. (b) Evaluating the consistency of knowledge learned among clients from a global view, namely, the CKA similarity of different local models at the same layer (feature extractor).} 
\label{fig10}
\end{figure*}

\subsubsection{Qualitative Analysis of Knowledge Preservation.} 

This section presents a qualitative analysis of knowledge preservation and examines the workings of FedCT from both local and global perspectives. We assess local knowledge distillation (LKD) by measuring the CKA representation similarity before and after cross-training on the same client. For global knowledge distillation (GKD), we compare CKA similarities between models from different clients. As shown in Figure~\ref{fig10}, the fused knowledge distillation (FKD) method, which integrates both global and local knowledge, achieves the highest test accuracy. Specifically, LKD yields the best local representation retention, followed by FKD, while GKD performs the worst (Figure~\ref{fig10}(a)). This aligns with expectations, as LKD directly addresses representation drift caused by model exchange. In contrast, GKD promotes consistent global representations by using a shared knowledge repository across clients. Although FKD does not achieve the best CKA similarity in either local or global view, it leads to superior test accuracy. This highlights the importance of expanding the effective training sample set for local models while maintaining inter-client knowledge consistency, and underscores the need to jointly consider both local and global factors in federated learning.


\subsubsection{Error Analysis of FedCT.}

\begin{figure*}[t]
\centering
\includegraphics[width=0.9\textwidth]{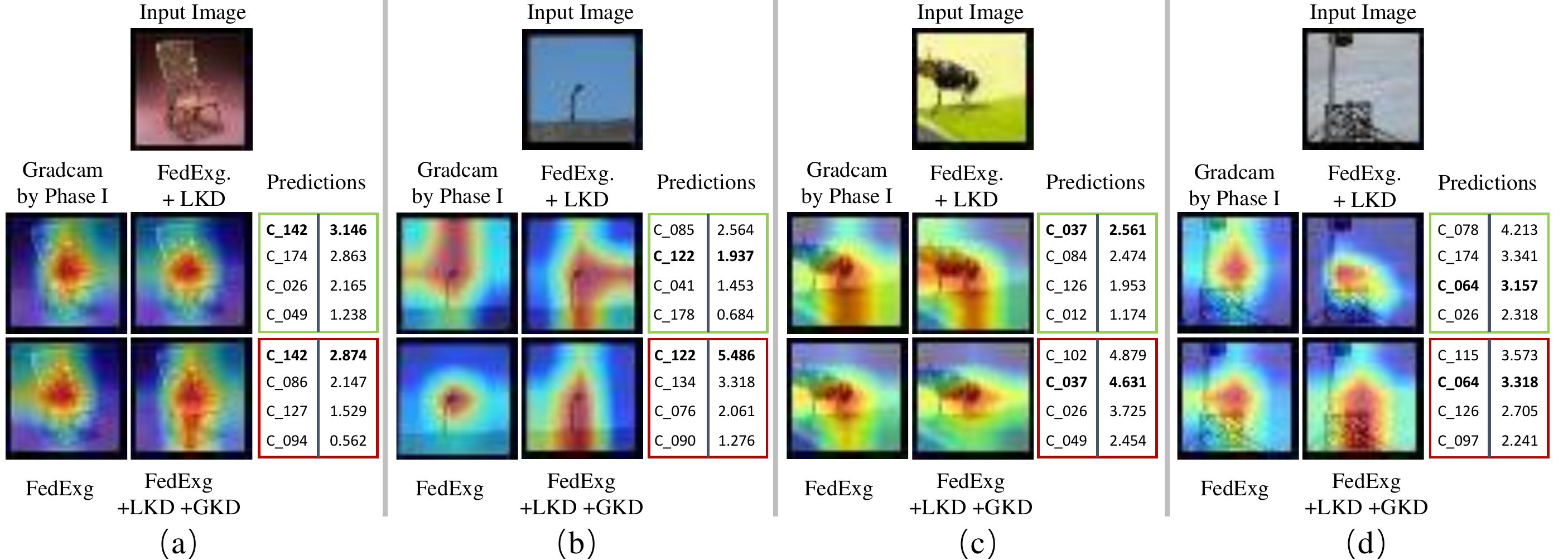}
   \caption{Error analysis of FedCT. (a) Cross-training can reinforce the knowledge acquired in the Phase 1. (b) FedCT is capable of leveraging global knowledge to assist the model in rectifying errors from the Phase 1. (c) FedCT fails due to poor data quality. (d) FedCT minimizes the prediction discrepancy between the ground truth and top-1.} 
\label{fig6}
\end{figure*}

This section delves into the working of FedCT, with a specific emphasis on feature attention and model outputs. Additionally, we employ GradCAM \cite{selvaraju2017grad} to generate attention maps. As depicted in Figure \ref{fig6}(a), the model in the first stage is able to focus on the target features. It is worth noting that FedExg, PGCT, and FedCT can all acquire new knowledge during the cross-training phase, which enables them to consistently broaden and reinforce their decision boundaries by harnessing data from diverse sources. Therefore, this further strengthens the predictions for “chair”. Figure \ref{fig6}(b) illustrates that when the model in the first phase has difficulty capturing the main features, FedExg fails due to knowledge forgetting, PGCT makes incorrect predictions with the guidence of unreliable knowledge, while FedCT can correct errors and compensate for shortcomings using global knowledge. In Figure \ref{fig6}(c), it can be observed that the model exhibits poor attention during the first stage. FedCT does not correct its prediction by cross-training, while PGCT makes the correct decision but seems uncertain. In Figure \ref{fig6}(d), both methods make errors. Especially, FedCT shows better attention to the key region and reduces the prediction discrepancy between ground-truth and the top-1 prediction. These observations provide further validation of the effectiveness of federated classification via cross-training. Furthermore, the balance between global and local knowledge is also highly crucial.



\section{Conclusion}
This paper proposes a multi-view knowledge-guided cross-training approach, termed FedCT, to tackle the issue of knowledge forgetting. Specifically, FedCT first introduces a consistency-aware knowledge broadcasting strategy, which is capable of enhancing collaboration between clients. Subsequently, FedCT develops a multi-view knowledge-guided representation learning method to guide client learn comprehensive knowledge, which extends the local training set and builds a unified knowledge base across clients as comprehensively as possible. And FedCT then leverages mixup-based feature augmentation to aggregate information, which increases the diversity of the feature space. Experimental results show that FedCT is capable of effectively mitigating knowledge forgetting from both local and global views, which leads to FedCT outperforming existing methods.

Despite FedCT has achieved impressive improvements, there are still two directions that can be further explored in future work. First, stronger local representation learning methods can guide feature alignment across clients, which can significantly enhance performance. Second, the investigation into more optimal exchange strategies is imperative, as it holds the potential to expedite model convergence.

\balance
\bibliographystyle{IEEEtran}

\bibliography{transbib}

\newpage

 




\vfill

\end{document}